\documentclass[a4paper]{article}

\usepackage{INTERSPEECH2022}
\usepackage{lipsum}
\usepackage{url}

\title{Deep LSTM Spoken Term Detection using Wav2Vec 2.0 Recognizer}
\name{Jan Švec, Jan Lehečka, Luboš Šmídl}
\address{
  Department of Cybernetics, University of West Bohemia, Pilsen, Czech Republic}
\email{\{honzas,jlehecka,smidl\}@kky.zcu.cz }

\begin{document}

\maketitle
\begin{abstract}
In recent years, the standard hybrid DNN-HMM speech recognizers are outperformed by the end-to-end speech recognition systems. One of the very promising approaches is the grapheme Wav2Vec 2.0 model, which uses the self-supervised pretraining approach combined with transfer learning of the fine-tuned speech recognizer. Since it lacks the pronunciation vocabulary and language model, the approach is suitable for tasks where obtaining such models is not easy or almost impossible.

In this paper, we use the Wav2Vec speech recognizer in the task of spoken term detection over a large set of spoken documents. The method employs a deep LSTM network which maps the recognized hypothesis and the searched term into a shared pronunciation embedding space in which the term occurrences and the assigned scores are easily computed.

The paper describes a bootstrapping approach that allows the transfer of the knowledge contained in traditional pronunciation vocabulary of DNN-HMM hybrid ASR into the context of grapheme-based Wav2Vec. The proposed method outperforms the previously published system based on the combination of the DNN-HMM hybrid ASR and phoneme recognizer by a large margin on the MALACH data in both English and Czech languages.
\end{abstract}

\section{Introduction}

The spoken term detection (STD) task is a widely studied field of speech processing. The STD emerged as a variant of traditional keyword spotting which speeds up the search phase by {offline} pre-processing and indexing of the searched data \cite{karakos2015}, where the pre-processing costs are counterweighted by the speed of an online search. A conventional approach to STD is to use the DNN-HMM hybrid speech recognizer to transform the input audio data into a set of word lattices from which the inverted word index is built. The drawback of this approach is the inability to index the out-of-vocabulary words (OOVs), which must be handled by other methods, such as the use of proxy-words  \cite{Zhiqiang2017,Chen2013} or sub-word units \cite{Psutka2011,he2016,heerden2017}. The OOVs are problematic especially in the domain of oral history archives processing, where the OOVs mostly represent the valuable searched terms, such as personal names and geographical terms \cite{gustman2002}.

Historically, the speech recognizers used for oral history archives included a large amount of domain-specific human-made knowledge incorporated into the pronunciation vocabulary or language model \cite{chylek2019}. Despite the advances in DNN-based acoustic modeling the hybrid speech recognition approach reached its limits and is overcome by the neural end-to-end systems. Especially for the oral history archives, the Wav2Vec 2.0 speech recognition approach \cite{wav2vec} is very promising. The method uses no language model nor vocabulary, which makes it ideal for modeling the OOVs. On the other hand, the knowledge accumulated in pronunciation vocabularies and language models is very valuable and could be exploited by the trainable end-to-end STD system.

\begin{figure}[t]
  \centering
  \includegraphics[width=1\linewidth]{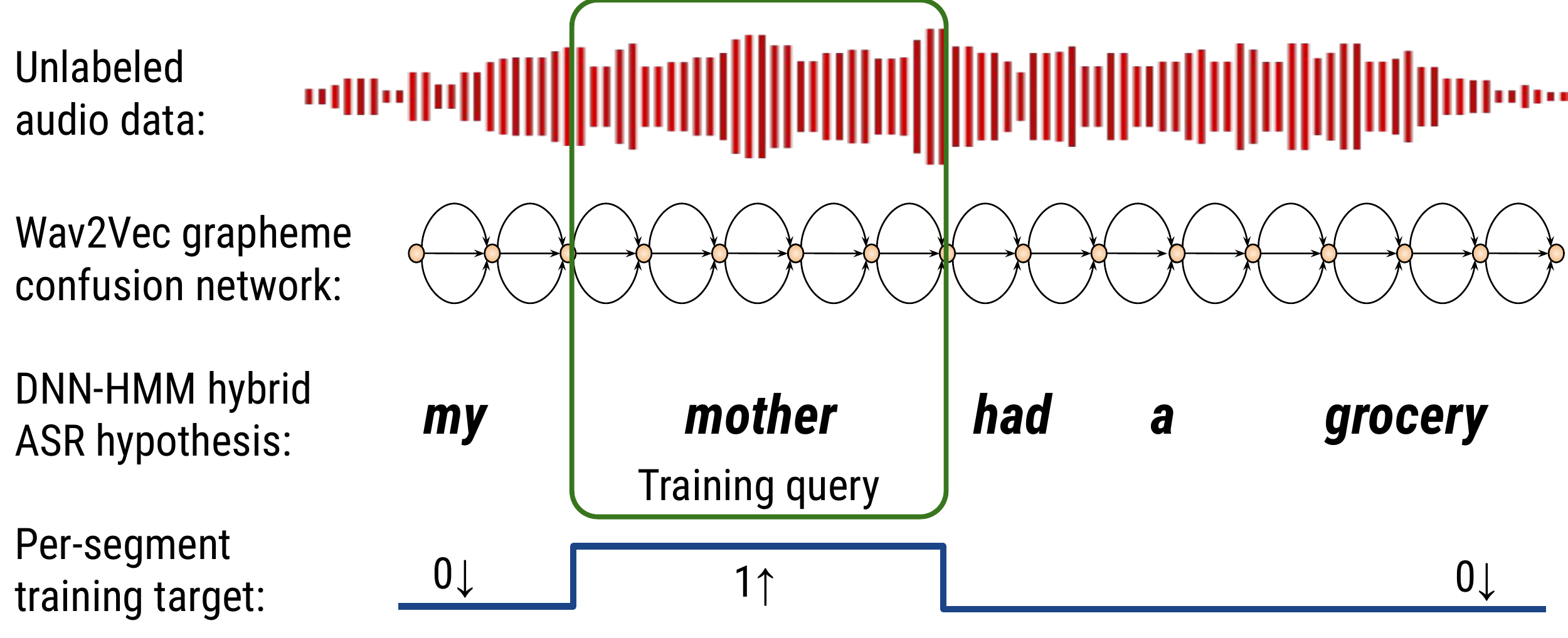}
  \caption{Schema of training query extraction from unlabeled audio data.}
  \label{fig:training}
  \vspace{-1.5em}
\end{figure}

The use of trainable models in the STD task or related keyword spotting and query-by-example tasks is widely studied. For example, the approach of \cite{Wang2018} maps input utterance to a sequence of vectors for subsequent STD, authors of \cite{yusuf21_interspeech} train the projection from acoustic features to an embedding space shared with the projected embedding of the query. A similar approach was also used in \cite{Sacchi2019}.

In this paper, we propose a method which combines the Wav2Vec and the DNN-HMM speech recognizers. First, the input audio data is recognized using the grapheme Wav2Vec recognizer to the form of grapheme confusion networks. Since the Wav2Vec recognizer does not use the pronunciation vocabulary, it produces an orthographic transcription of the utterance. Then, the same data are recognized using the traditional DNN-HMM hybrid recognizer and the high-confidence recognized words are then used as query terms and the corresponding segments of the grapheme confusion network are used as samples of the particular occurrence of the query term (Fig. \ref{fig:training}). The query and the corresponding training targets (binary value indicating the occurrence of the query in the audio) are used to train the Deep LSTM STD neural network \cite{svec2021}, which maps the pair into a joint embedding space in which the score of the putative hit is computed. This way, the Deep LSTM STD network inherently learns the phonetic and syntactic knowledge incorporated in the DNN-HMM hybrid recognizer.

\section{Wav2Vec pretraining \& fine-tuning}

The Wav2Vec 2.0 framework \cite{wav2vec} is a self-supervised framework for learning representations from raw audio data. It uses a multi-layer convolutional neural network to compute frame-level features. The features are processed using a Transformer network to predict the context-dependent representation of the input audio. After pre-training on unlabeled speech, the dense classification layer with softmax activation is added on top of the Transformer and trained in a supervised manner using the CTC loss \cite{baevski2020ctc} to obtain the grapheme-level speech recognizer.

\subsection{Grapheme confusion network}
\label{sec:gcn}

The output of the CTC classification layer on top of the Wav2Vec is virtually a matrix with rows corresponding to the frame index axis of the input signal and columns are posterior probabilities of the output symbols in the given input frame. In other words, the output is as a sequence of grapheme posterior probabilities which could be easily decoded into a single grapheme-level hypothesis following the CTC approach: (1)~compute the most probable symbol for each time frame; (2)~condense the consecutive occurrences of the same symbol into a single output symbol and (3)~drop all \emph{blank} symbols. 

The CTC classification head is often trained to also output the special symbol used as word separator (denoted as $\mid$). In the experiments, we merged this symbol with the blank symbol, which we will denote as $\epsilon$ in the following paragraphs. All other symbols in the output dictionary represent the graphemes of the given language.

We verified, that although the CTC training considers all output sequences with the same decoded graphemes and different alignment as equivalent, the grapheme-level alignments obtained from the above-mentioned decoding are sufficient to effectively generate the frame-grapheme correspondence. In other words, the grapheme-level Wav2Vec produces time alignments that could be used in the STD task (see Tab. \ref{tab:lvcsr}).

To exploit the alternative hypotheses beyond the 1-best hypothesis from the CTC classifier, we designed an algorithm that converts the output matrix of grapheme posteriors into a grapheme confusion network with frame-aligned nodes. 
The input of the algorithm is the matrix of posterior probabilities $\vec{P} = [p_{ts}]$
where the axis $t=1,\ldots T$ corresponds to an index of an input frame and the axis of output graphemes $s \in V$ represents a vector of posterior probabilities $\vec{p}_t$ defined over the output vocabulary $V$. We assume that there is a fixed bijective mapping from the symbols $s$ to the columns of the matrix $\vec{P}$. Therefore, we will use the symbol $s$ directly as index in the following text for simplicity.

The algorithm first generates the 1-best grapheme hypothesis of $N$ graphemes $H = (h_i)_{i=1}^N, h_i \in V$ and the corresponding frame-alignment $A = (b_i, e_i)_{i=1}^N$ where $1 \leq b_i < e_i\leq T+1$ and $e_i \leq b_{i+1}$. The integer numbers $b_i$ and $e_i$ define the first and last frame of the $i$-th grapheme of the hypothesis $H$.
The 1-best hypothesis $H$ and the frame-alignment $A$ satisfy the following conditions:

\vspace{-1em}
\begin{equation}
\begin{split}
    h_i &= \arg\max_{s} p_{ts} ~~~ \forall t \in [b_i, d_i);~~b_i < d_i \le e_i \\
    \epsilon &= \arg\max_{s} p_{ts} ~~~ \forall t \in [d_i, e_i)
\end{split}
\end{equation}

\noindent In other words\footnote{Here we use the interval notation also for the integer intervals: $[b_i, d_i)$ means integers $b_i, b_i+1, \ldots, d_i-1$ }, the output grapheme $h_i$ is the most probable predicted symbol on the frames spanning indices $[b_i, d_i)$ followed by zero or more frames with indices $[d_i, e_i)$ which are mapped to the blank $\epsilon$ symbol in the vocabulary $V$ and are used to separate two subsequent occurrences of the same grapheme.

\noindent \textbf{Example:} The word \emph{book} could be decoded as per-frame most probable sequence of the following symbols: \emph{b-b-$\epsilon$-o-o-$\epsilon$-o-k-k}. Then, the sequence of 1-best grapheme hypothesis is $H = (b, o, o, k)$ and the corresponding alignment $A = \left((1, 4), (4, 7), (7, 8), (8, 10)\right) $.

The 1-best grapheme hypothesis $H$ and the alignment $A$ is used as pivot sequence in confusion network generation \cite{HakkaniTur2006}, i.e. the alternative graphemes are added parallel for each grapheme $h_i$. The $i$-th segment of the confusion network could be described as a probability distribution $\vec{c}_i = [c_{is}]$ of symbols $s \in \bar{V}$ defined over the vocabulary with the $\epsilon$ symbol removed, i.e. $\bar{V} = V \setminus \{\epsilon\}$. In other words, the probability mass of the blank symbol is distributed across the graphemes in $\bar{V}$:

\begin{equation}
    c_{is} = \frac{\sum_{t \in [b_i, e_i)} p_{ts}}{\sum_{t \in [b_i, e_i), \bar{s} \in \bar{V} } p_{t\bar{s}}}
\end{equation}

The resulting grapheme confusion network is represented by the sequence of segments $C = (\vec{c}_i)_{i=1}^N$ representing the average grapheme posterior probabilities computed over the corresponding frame spans $A = (b_i, e_i)_{i=1}^N$. 

We have to note that the CTC loss does not guarantee that the output grapheme is generated for the first frame the grapheme occurs. We address such CTC alignment inaccuracies in the training process described in Sec. \ref{sec:deeplstm}.

\subsection{Sliding window CTC decoding}

The Wav2Vec model is an architecture based on the Transformer model. The self-attention layers present in the Transformer blocks impose a quadratic computational and memory complexity with respect to the input sequence length. Therefore a manageable length of input waveforms is in the order of tens of seconds. To overcome this limitation, we used the sliding window approach to obtain the grapheme posteriors $\vec{P}$. In particular, we used the window length of 18 seconds with a 3-second overlap. The overlaps are split in half and the posteriors for the first 1.5 seconds are taken from the left window and posteriors for the remaining 1.5 seconds from the right window. This way, we are able to compute the posteriors $\vec{P}$ for inputs of arbitrary lengths. The grapheme confusion network is then computed for the whole input based on the generated matrix $\vec{P}$.

\section{Deep LSTM Spoken Term Detection}
\label{sec:deeplstm}

\begin{figure}[t]
  \centering
  \includegraphics[width=1\linewidth]{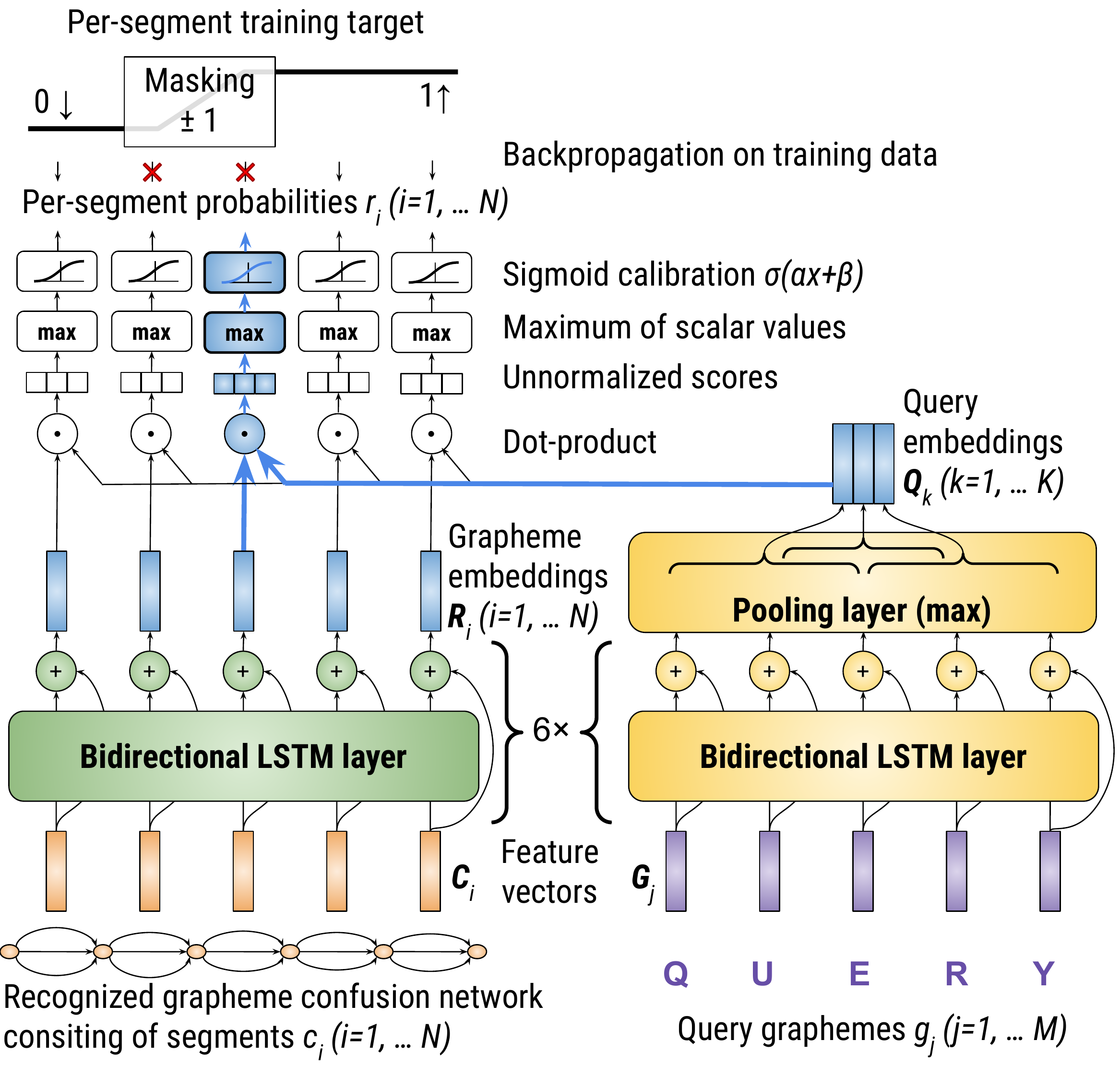}
  \caption{Deep LSTM network architecture. In blue color, the scoring for a single confusion network segment is highlighted (see Eq. (\ref{eq:prob}) for details). The network for estimation of the minimal number of segments for the query is omitted for clarity.}
  \label{fig:deeplstm}
  \vspace{-1.5em}
\end{figure}

The overall network consists of two independent processing pipelines implemented using deep LSTM networks: (1) the recognition output projection and (2) the searched term projection and minimum length estimation (Fig. \ref{fig:deeplstm}). In each pipeline, we use a stack of 6 bidirectional LSTM layers with skip connections and dimensionality 300 (concatenated forward and backward LSTM outputs) for both processing pipelines. 

To process the recognition output in the form of a grapheme confusion network we use the features computed the same way as in \cite{svec2021} from the segments of the confusion network $\vec{c}_i$: time duration of the $i$-th segment ($e_i-b_i$), the grapheme $s$  and its probability $c_{is}$ for the 3 most probable graphemes.

The confusion network segments $\vec{c}_i$ are transformed into a sequence of feature vectors $\vec{C}_i$, which are mapped using a stack of LSTM network with skip connections (confusion network processing pipeline, green blocks in Fig. \ref{fig:deeplstm}) to the sequence of embedding vectors $\vec{R}_i$. The vectors $\vec{R}_i$ are independent of the query and could be pre-computed ahead of time to speed up the STD.

The graphemes of the query ($g_j, j=1, \ldots M$) are mapped using an input embedding layer to features $\vec{G}_j$. The same network architecture followed by the maximum pooling layer (grapheme processing pipeline, yellow blocks in Fig. \ref{fig:deeplstm}) is used to produce a sequence of vectors $\vec{Q}_k, k=1, \ldots K$. 

The output, calibrated probability $r_i, i=1, \ldots N$ of the occurrence is computed as:

\begin{equation}
    r_i = \sigma\left(\alpha \cdot \max_{k=1}^K ( \vec{R}_i \cdot \vec{Q}_k ) + \beta\right)
    \label{eq:prob}
\end{equation}

\noindent where $\sigma(x)=\frac{1}{1+e^{-x}}$ denotes the sigmoid function and $\alpha$ and $\beta$ are trainable calibration parameters.

The pooling layer of the grapheme processing pipeline allows extracting the query embeddings $\vec{Q}_k$ for different parts of the query word. We use more than one vector $\vec{Q}_k$, because it increases the expressive power of the network without increasing the number of trainable parameters. Based on \cite{svec2021}, we use $K=3$. Then $\vec{Q}_1$ represents the first half of the query, $\vec{Q}_2$ the middle of the query, and $\vec{Q}_3$ the second half. 

The sequence of grapheme embeddings $\vec{G}_i$ is also used to predict the expected minimum number of confusion network segments for the query. For this part of the network, we use a simple bidirectional LSTM with 20 units for each direction, and the final output is computed using a dense linear layer.


As we have said in Sec. \ref{sec:gcn}, the output alignment $A=(b_i, e_i)$ and the corresponding grapheme confusion network segments $\vec{c}_i$ are not precisely aligned with the input acoustic signal due to the use of the CTC loss during the training. We analyzed the training queries generated from high-confidence DNN-HMM hybrid recognizer outputs, and we observed that the mismatch between the recognized word and corresponding 1-best graphemes from the Wav2Vec recognizer is at most a few graphemes (and a few corresponding confusion network segments). 
We did not observe the case where the CTC output was shifted by more than five graphemes. Therefore, we incorporated this uncertainty into the training process by simple masking of the network output. We say, that we use masking $\pm n$ if the back-propagation of the gradient during network training on the $n$ confusion network segments before and after the left and right boundary of the training query are ignored. In other words, if there is a 0-to-1 or 1-to-0 transition in the training target, then the loss of $2n$ surrounding segments is not back-propagated, and the network itself decides what is the best matching "soft transition" (illustrated in the top part of Fig. \ref{fig:deeplstm}).


In prediction time, the output from the Wav2Vec is converted to the grapheme confusion network and subsequently projected using the confusion network pipeline into a sequence of embeddings $\vec{R}_i$. The arbitrary query represented as a sequence of graphemes $g_j$ is converted to a sequence of embedding $\vec{Q}_k$. The probabilities $r_i$ are computed according to Eq.~\ref{eq:prob}. We use simple thresholding to detect peaks in $r_i$ of at least the minimum length estimated from $g_j$. The score of a hit is computed as a mean value of the peak and the time boundaries of the hit are computed from an alignment $(b_i, e_i)$. More details could be found in \cite{svec2021}.

\section{Dataset \& model description}

The presented method was evaluated on the data from a USC-SFI MALACH archive in two languages -- English \cite{MALACHen} and Czech \cite{MALACHcz}. The archive for each language was recognized using the DNN-HMM hybrid ASR; basic statistics are summarized in Tab. \ref{tab:stats} (more details are provided in \cite{svec2017}). 

For generating the grapheme confusion networks, we used the Wav2Vec model fine-tuned on the MALACH data. For English, we started with publicly available pretrained \texttt{Wav2Vec 2.0 Base} model\footnote{Downloaded from \url{https://dl.fbaipublicfiles.com/fairseq/wav2vec/wav2vec_small.pt}} and fine-tuned on the MALACH train data with the same setting as in \cite{wav2vec} using \texttt{Fairseq} tool\footnote{\url{https://github.com/pytorch/fairseq}}.
We sliced long training audio signals on speech pauses not to exceed the length of 30\,s. We removed non-speech events and punctuation from the transcripts and mapped all graphemes into lowercase. We fine-tuned the English model with a peak learning rate $8 \times 10^{-5}$ for 80 thousand updates, which corresponds approximately to 86 epochs over 246 hours of training audio data. For Czech, there is no public monolingual pretrained model, so we decided to train our own model\footnote{Available at \url{https://huggingface.co/fav-kky/wav2vec2-base-cs-80k-ClTRUS}}. We gathered more than 80 thousand hours of Czech speech data, including records from TV shows (27k hours), parliament (24k hours), radio (22k hours), sports (5k hours), telephone data (2k hours), and several other domains. We sliced all records not to exceed 30\,s and followed the same pretraining steps as for the base Wav2Vec 2.0 model in \cite{wav2vec}.
We prepared our fine-tuning data for the Czech model the same way as for English and fine-tuned with a peak learning rate $2 \times 10^{-5}$ for 80 thousand updates, which corresponds approximately to 270 epochs over 87 hours of training audio data.
The CTC classification layer predicts probabilities of 53 symbols for English and 51 for Czech, and the output frame length is 0.02s for both models. 

The in- and out-of-vocabulary terms were selected automatically from the development and test data based on the DNN-HMM recognition vocabulary. We filtered all possible terms so that the terms are not substrings of other terms in the dataset nor the words in the vocabulary. The numbers reported in this paper are directly comparable to results presented in \cite{svec2017,svec2021}.

\begin{table}[t]
    \small
  \caption{Statistics of development and test sets \cite{svec2017}. ASR means DNN-HMM hybrid ASR.}
  \label{tab:stats}
  \centering
  \begin{tabular}{lrrrr}
    \toprule[0.9pt]
    & \multicolumn{2}{c}{English} & \multicolumn{2}{c}{Czech} \\
    \cmidrule(l){2-3} \cmidrule(l){4-5}
    & Dev & Test & Dev & Test \\
    \cmidrule(l){1-5}
    ASR vocabulary size & \multicolumn{2}{c}{243,699} & \multicolumn{2}{c}{252,082} \\
    \#speakers  & 10 & 10 & 10 & 10  \\
    OOV rate & 0.5\% & 3.2\% & 0.3\% & 2.6\%  \\
    ASR word error rate & 24.10 & 19.66 & 23.98 & 19.11 \\
    \#IV terms & 597 & 601 & 1680 & 1673 \\
    \#OOV terms & 31 & 6& 1145 & 948 \\
    dataset length $[$hours$]$ & 11.1 & 11.3 & 20.4 & 19.4 \\
    \bottomrule[0.9pt]
  \end{tabular}
\end{table}


\section{Experimental evaluation}

In the following experiments, we report the actual term weighted value (ATWV) metric evaluated on the test data  \cite{Computer2013}. The optimal decision threshold was determined on the development data and therefore the Tab. \ref{tab:lvcsr} and \ref{tab:results} present the maximum TWV (MTWV) on this data.

The first experiment was focused on the comparison of the grapheme Wav2Vec models with respect to the DNN-HMM hybrid ASR as a baseline. In this experiment, we focused on the STD performance evaluated on the in-vocabulary terms only to obtain comparable outcomes. We also performed the CTC decoding using beam search and language model\footnote{We used an implementation from Huggingface Transformers \texttt{Wav2Vec2ProcessorWithLM} \cite{wolf2020transformers}.}. In this experiment, we used the same language model as in the DNN-HMM hybrid ASR. The results are compared in Tab. \ref{tab:lvcsr}. The table shows that the raw grapheme Wav2Vec recognizer performance is worse than the DNN-HMM baseline. The addition of a language model (without the pronunciation vocabulary) improves the STD performance, it reaches or exceeds the performance of the baseline.

\begin{table}[t]
    \small
  \vspace{-0.5em}
  \caption{In-vocabulary terms, results on the development dataset (MTWV).}
  \label{tab:lvcsr}
  \centering
  \begin{tabular}{lrr}
    \cmidrule[0.9pt](l){1-3}
               & English & Czech \\
    \cmidrule(l){1-3}
    DNN-HMM hybrid ASR                & 0.7899 & 0.8760 \\
    Wav2Vec 2.0 grapheme recognizer  & 0.6178 & 0.7248 \\
    Wav2Vec 2.0 LM + beam search      & 0.8105 & 0.8757 \\
    \cmidrule[0.9pt](l){1-3}
  \end{tabular}
  \vspace{-0.5em}
\end{table}

Although the performance on in-vocabulary terms could be further improved by the addition of pronunciations, we focused on the experiments with a more general method, which uses the trainable Deep LSTM STD framework described in this paper. As we have described, the baseline DNN-HMM hybrid ASR is used just for the generation of training queries, and the mapping from normalized to orthographic transcription of the query is trained in this framework for both the in-vocabulary and the out-of-vocabulary terms.

We trained the described Deep LSTM STD neural networks for each language using the unlabeled English and Czech data from the Visual History Archive. The recordings of speakers occurring in development/test data were not used in training. The models were trained for 200k steps, each batch consisted of 32 samples. We used an ADAM optimizer and inverse square root learning rate schedule without warm-up. To generate the training queries, only the words with a confidence score higher than 0.95 were used. 

To determine the optimum width of the output masking we chose the value which optimizes the MTWV metric on development data. The optimum width was $\pm 1$ for English and $\pm 2$ for Czech. The Deep LSTM STD outperforms the baseline methods by a large margin. The proposed method also provides better results than the same network architecture based on the DNN-HMM phoneme recognizer \cite{svec2021}. The last two rows of Tab.~\ref{tab:results} show the evaluation on in-vocabulary and out-of-vocabulary terms only. The method provides better results on in-vocabulary terms than the Wav2Vec recognizer with the language model and beam search and at the same time only slightly worse results for the out-of-vocabulary terms, which are generally impossible to find by the vocabulary-based recognizers.

The results evaluated on test data are shown in Tab. \ref{tab:results-test}. The table shows the progress of the STD methods from the baseline empirical one through the use of Siamese neural networks for scoring and ending with the fully trainable Deep LSTM STD framework. If we use the Wav2Vec grapheme recognizer generating the grapheme confusion networks, it outperforms the same method using the DNN-HMM baseline even if we boost the performance using the ASR output (LM and vocabulary) to search for in-vocabulary words.

\begin{table}[t]
    \small
  \caption{Results on the development dataset (MTWV) using grapheme-based STD.}
  \label{tab:results}
  \centering
  \begin{tabular}{lrr}
    \cmidrule[0.9pt](l){1-3}
     & English & Czech \\
    \cmidrule(l){1-3}
     Empirical method \cite{Psutka2011}   & 0.4636 & 0.6225 \\
     Siamese neural network \cite{svec2017} & 0.5012 & 0.6547 \\
    \cmidrule(l){1-3}
     Deep LSTM / DNN-HMM \cite{svec2021} & 0.6703 & 0.7723 \\
    \cmidrule(l){1-3}
     \multicolumn{3}{l}{\textbf{Deep LSTM / Wav2Vec 2.0} (proposed method)} \\
     ~~ no masking         & 0.8198 & 0.8823 \\
     ~~ masking $\pm 1$    & \textbf{0.8308} & 0.8888 \\
     ~~ masking $\pm 2$    & 0.7873 & \textbf{0.8987} \\
     ~~ masking $\pm 3$    & 0.7293 & 0.8790 \\
    \cmidrule(l){1-3}
    ~~ In-vocabulary terms      & 0.8398 & 0.9117 \\
    ~~ Out-of-vocabulary terms  & 0.7027 & 0.8784 \\
    \cmidrule[0.9pt](l){1-3}
  \end{tabular}
\end{table}

\begin{table}[t]
    \small
  \vspace{-0.5em}
  \caption{Results on the test dataset (ATWV).}
  \label{tab:results-test}
  \centering
  \begin{tabular}{lrr}
    \cmidrule[0.9pt](l){1-3}
               & English & Czech \\
    \cmidrule(l){1-3}
    Empirical method \cite{Psutka2011}   & 0.4435 & 0.6564 \\
    Siamese neural network \cite{svec2017}  & 0.4956 & 0.6873 \\
    \cmidrule(l){1-3}
    Deep LSTM / DNN-HMM \cite{svec2021} & 0.5823 & 0.7531 \\
    ~~ with LM and vocabulary & 0.7275 & 0.8283 \\
    \cmidrule(l){1-3}
    \textbf{Deep LSTM / Wav2Vec 2.0} & \textbf{0.7616} & \textbf{0.9100} \\
    \cmidrule[0.9pt](l){1-3}
  \end{tabular}
  \vspace{-0.5em}
\end{table}

\section{Conclusions}

In this paper, we presented the fully trainable spoken term detection framework based on deep LSTMs and Wav2Vec recognizer. The output of Wav2Vec speech recognizer was converted into the grapheme confusion network using the sliding window CTC decoding approach to be able to handle inputs of arbitrary length. The DNN-HMM hybrid ASR was used to generate STD training queries by using blindly recognized unlabeled speech data. We were able to train an end-to-end STD system, which outperforms not only the DNN-HMM baseline but also the Wav2Vec grapheme recognizer and also the language model enhanced Wav2Vec. At the same time, the STD system allows searching for out-of-vocabulary terms. 

\section{Acknowledgements}

This research was supported by the Ministry of the Interior of the Czech Republic, project No. VJ01010108. Computational resources were supplied by the project "e-Infrastruktura CZ" (e-INFRA LM2018140).

\bibliographystyle{IEEEtran}

\newpage

\bibliography{mybib}

\end{document}